# Combining Knowledge- and Corpus-based Word-Sense-Disambiguation Methods

**Andres Montoyo**                                    MONTOYO@DLSI.UA.ES
*Dept. of Software and Computing Systems*
*University of Alicante, Spain*

**Armando Suárez**                                    ARMANDO@DLSI.UA.ES
*Dept. of Software and Computing Systems*
*University of Alicante, Spain*

**German Rigau**                                      RIGAU@SI.EHU.ES
*IXA Research Group*
*Computer Science Department*
*Basque Country University, Donostia*

**Manuel Palomar**                                    MPALOMAR@DLSI.UA.ES
*Dept. of Software and Computing Systems*
*University of Alicante, Spain*

## Abstract

In this paper we concentrate on the resolution of the lexical ambiguity that arises when a given word has several different meanings. This specific task is commonly referred to as word sense disambiguation (WSD). The task of WSD consists of assigning the correct sense to words using an electronic dictionary as the source of word definitions. We present two WSD methods based on two main methodological approaches in this research area: a knowledge-based method and a corpus-based method. Our hypothesis is that word-sense disambiguation requires several knowledge sources in order to solve the semantic ambiguity of the words. These sources can be of different kinds— for example, syntagmatic, paradigmatic or statistical information. Our approach combines various sources of knowledge, through combinations of the two WSD methods mentioned above. Mainly, the paper concentrates on how to combine these methods and sources of information in order to achieve good results in the disambiguation. Finally, this paper presents a comprehensive study and experimental work on evaluation of the methods and their combinations.

## 1. Introduction

Knowledge technologies aim to provide meaning to the petabytes of information content that our multilingual societies will generate in the near future. Specifically, a wide range of advanced techniques are required to progressively automate the knowledge lifecycle. These include analyzing, and then automatically representing and managing, high-level meanings from large collections of content data. However, to be able to build the next generation of intelligent open-domain knowledge application systems, we need to deal with concepts rather than words.

## 1.1 Dealing with Word Senses

In natural language processing (NLP), word sense disambiguation (WSD) is defined as the task of assigning the appropriate meaning (sense) to a given word in a text or discourse. As an example, consider the following three sentences:

1. Many cruise missiles have **fallen** on Baghdad.

2. Music sales will **fall** by up to 15% this year.

3. U.S. officials expected Basra to **fall** early.

Any system that tries to determine the meanings of the three sentences will need to represent somehow three different senses for the verb *fall*. In the first sentence, the *missiles have been launched on Baghdad*. In the second sentence, *sales will decrease*, and in the third *the city will surrender early*. WordNet 2.0 (Miller, 1995; Fellbaum, 1998)[1] contains thirty-two different senses for the verb *fall* as well as twelve different senses for the noun *fall*. Note also that the first and third sentence belong to the same, military domain, but use the verb *fall* with two different meanings.

Thus, a WSD system must be able to assign the correct sense of a given word, in these examples, *fall*, depending on the context in which the word occurs. In the example sentences, these are, respectively, senses 1, 2 and 9, as listed below.

- 1. fall—"descend in free fall under the influence of gravity" ("The branch fell from the tree"; "The unfortunate hiker fell into a crevasse").

- 2. descend, fall, go down, come down—"move downward but not necessarily all the way" ("The temperature is going down"; "The barometer is falling"; "Real estate prices are coming down").

- 9. fall—"be captured" ("The cities fell to the enemy").

Providing innovative technology to solve this problem will be one of the main challenges in language engineering to access advanced knowledge technology systems.

## 1.2 Word-Sense Disambiguation

Word sense ambiguity is a central problem for many established Human Language Technology applications (e.g., machine translation, information extraction, question answering, information retrieval, text classification, and text summarization) (Ide & Véronis, 1998). This is also the case for associated subtasks (e.g., reference resolution, acquisition of subcategorization patterns, parsing, and, obviously, semantic interpretation). For this reason, many international research groups are working on WSD, using a wide range of approaches. However, to date, no large-scale, broad-coverage, accurate WSD system has been built (Snyder & Palmer, 2004). With current state-of-the-art accuracy in the range 60–70%, WSD is one of the most important open problems in NLP.

---

1. http://www.cogsci.princeton.edu/~wn/

Even though most of the techniques for WSD usually are presented as stand-alone techniques, it is our belief, following McRoy (1992), that full-fledged lexical ambiguity resolution will require to integrate several information sources and techniques.

In this paper, we present two complementary WSD methods based on two different methodological approaches, a *knowledge-based* and a *corpus-based* methods, as well as several methods that combine both into hybrid approaches.

The knowledge-based method disambiguates nouns by matching context with information from a prescribed knowledge source. WordNet is used because it combines the characteristics of both a dictionary and a structured semantic network, providing definitions for the different senses of the English words and defining groups of synonymous words by means of *synsets*, which represent distinct lexical concepts. WordNet also organizes words into a conceptual structure by representing a number of semantic relationships (hyponymy, hypernymy, meronymy, etc.) among synsets.

The corpus-based method implements a supervised machine-learning (ML) algorithm that learns from annotated sense examples. The corpus-based system usually represents linguistic information for the context of each sentence (e.g., usage of an ambiguous word) in the form of feature vectors. These features may be of a distinct nature: word collocations, part-of-speech labels, keywords, topic and domain information, grammatical relationships, etc. Based on these two approaches, the main objectives of the work presented in this paper are:

- To study the performance of different mechanisms of combining information sources by using *knowledge-based* and *corpus-based* WSD methods together.

- To show that a *knowledge-based* method can help a *corpus-based* method to better perform the disambiguation process and vice versa.

- To show that the combination of both approaches outperforms each of the methods taken individually, demonstrating that the two approaches can play complementary roles.

- Finally, to show that both approaches can be applied in several languages. In particular, we will perform several experiments in Spanish and English.

In the following section a summary of the background of word sense disambiguation is presented. Sections 2.1 and 2.2 describe the knowledge-based and corpus-based systems used in this work. Section 3 describes two WSD methods: the specification marks method and the maximum entropy-based method. Section 4 presents an evaluation of our results using different system combinations. Finally, some conclusions are presented, along with a brief discussion of work in progress.

## 2. Some Background on WSD

Since the 1950s, many approaches have been proposed for assigning senses to words in context, although early attempts only served as models for toy systems. Currently, there are two main methodological approaches in this area: knowledge-based and corpus-based methods. Knowledge-based methods use external knowledge resources, which define explicit sense distinctions for assigning the correct sense of a word in context. Corpus-based methods use machine-learning techniques to induce models of word usages from large collections of text examples. Both knowledge-based and corpus-based methods present different benefits and drawbacks.

## 2.1 Knowledge-based WSD

Work on WSD reached a turning point in the 1980s and 1990s when large-scale lexical resources such as dictionaries, thesauri, and corpora became widely available. The work done earlier on WSD was theoretically interesting but practical only in extremely limited domains. Since Lesk (1986), many researchers have used machine-readable dictionaries (MRDs) as a structured source of lexical knowledge to deal with WSD. These approaches, by exploiting the knowledge contained in the dictionaries, mainly seek to avoid the need for large amounts of training material. Agirre and Martinez (2001b) distinguish ten different types of information that can be useful for WSD. Most of them can be located in MRDs, and include part of speech, semantic word associations, syntactic cues, selectional preferences, and frequency of senses, among others.

In general, WSD techniques using pre-existing structured lexical knowledge resources differ in:

- the lexical resource used (monolingual and/or bilingual MRDs, thesauri, lexical knowledge base, etc.);

- the information contained in this resource, exploited by the method; and

- the property used to relate words and senses.

Lesk (1986) proposes a method for guessing the correct word sense by counting word overlaps between dictionary definitions of the words in the context of the ambiguous word. Cowie et al. (1992) uses the simulated annealing technique for overcoming the combinatorial explosion of the Lesk method. Wilks et al. (1993) use co-occurrence data extracted from an MRD to construct word-context vectors, and thus word-sense vectors, to perform a large set of experiments to test relatedness functions between words and vector-similarity functions.

Other approaches measure the relatedness between words, taking as a reference a structured semantic net. Thus, Sussna (1993) employs the notion of conceptual distance between network nodes in order to improve precision during document indexing. Agirre and Rigau (1996) present a method for the resolution of the lexical ambiguity of nouns using the Word-Net noun taxonomy and the notion of *conceptual density*. Rigau et al. (1997) combine a set of *knowledge-based* algorithms to accurately disambiguate definitions of MRDs. Mihalcea and Moldovan (1999) suggest a method that attempts to disambiguate all the nouns, verbs, adverbs, and adjectives in a given text by referring to the senses provided by Word-Net. Magnini et al. (2002) explore the role of domain information in WSD using WordNet domains (Magnini & Strapparava, 2000); in this case, the underlying hypothesis is that information provided by domain labels offers a natural way to establish semantic relations among word senses, which can be profitably used during the disambiguation process.

Although *knowledge-based* systems have been proven to be ready-to-use and scalable tools for all-words WSD because they do not require sense-annotated data (Montoyo et al., 2001), in general, supervised, *corpus-based* algorithms have obtained better precision than *knowledge-based* ones.

## 2.2 *Corpus-based* WSD

In the last fifteen years, empirical and statistical approaches have had a significantly increased impact on NLP. Of increasing interest are algorithms and techniques that come from the machine-learning (ML) community since these have been applied to a large variety of NLP tasks with remarkable success. The reader can find an excellent introduction to ML, and its relation to NLP, in the articles by Mitchell (1997), Manning and Schütze (1999), and Cardie and Mooney (1999), respectively. The types of NLP problems initially addressed by statistical and machine-learning techniques are those of language- ambiguity resolution, in which the correct interpretation should be selected from among a set of alternatives in a particular context (e.g., word-choice selection in speech recognition or machine translation, part-of-speech tagging, word-sense disambiguation, co-reference resolution, etc.). These techniques are particularly adequate for NLP because they can be regarded as classification problems, which have been studied extensively in the ML community. Regarding automatic WSD, one of the most successful approaches in the last ten years is supervised learning from examples, in which statistical or ML classification models are induced from semantically annotated corpora. Generally, supervised systems have obtained better results than unsupervised ones, a conclusion that is based on experimental work and international competitions[2]. This approach uses semantically annotated corpora to train machine–learning (ML) algorithms to decide which word sense to choose in which contexts. The words in such annotated corpora are tagged manually using semantic classes taken from a particular lexical semantic resource (most commonly WordNet). Many standard ML techniques have been tried, including Bayesian learning (Bruce & Wiebe, 1994), Maximum Entropy (Suárez & Palomar, 2002a), exemplar-based learning (Ng, 1997; Hoste et al., 2002), decision lists (Yarowsky, 1994; Agirre & Martinez, 2001a), neural networks (Towell & Voorhees, 1998), and, recently, margin-based classifiers like boosting (Escudero et al., 2000) and support vector machines (Cabezas et al., 2001).

Corpus-based methods are called "supervised" when they learn from previously sense-annotated data, and therefore they usually require a large amount of human intervention to annotate the training data (Ng, 1997). Although several attempts have been made (e.g., Leackock et al., 1998; Mihalcea & Moldovan, 1999; Cuadros et al., 2004), the knowledge acquisition bottleneck (too many languages, too many words, too many senses, too many examples per sense) is still an open problem that poses serious challenges to the supervised learning approach for WSD.

## 3. WSD Methods

In this section we present two WSD methods based, respectively, on the two main methodological approaches outlined above: a specification marks method (SM) (Montoyo & Palomar, 2001) as a knowledge-based method, and a maximum entropy-based method (ME) (Suárez & Palomar, 2002b) as a corpus-based method. The selected methods can be seen as representatives of both methodological approaches. The specification marks method is inspired by the conceptual density method (Agirre & Rigau, 1996) and the maximum entropy method has been also used in other WSD systems (Dang et al., 2002).

---

2. http://www.senseval.org

## 3.1 Specification Marks Method

The underlying hypothesis of this knowledge base method is that the higher the similarity between two words, the larger the amount of information shared by two of its concepts. In this case, the information commonly shared by several concepts is indicated by the most specific concept that subsumes them in the taxonomy.

The input for this WSD module is a group of nouns $W = \{w_1, w_2, ..., w_n\}$ in a context. Each word $w_i$ is sought in WordNet, each having an associated set of possible senses $S_i = \{S_{i1}, S_{i2}, ..., S_{in}\}$, and each sense having a set of concepts in the IS-A taxonomy (hypernymy/hyponymy relations). First, this method obtains the common concept to all the senses of the words that form the context. This concept is marked by the initial specification mark (ISM). If this initial specification mark does not resolve the ambiguity of the word, we then descend through the WordNet hierarchy, from one level to another, assigning new specification marks. For each specification mark, the number of concepts contained within the subhierarchy is then counted. The sense that corresponds to the specification mark with the highest number of words is the one chosen to be sense disambiguated within the given context. Figure 1 illustrates graphically how the word *plant*, having four different senses, is disambiguated in a context that also has the words *tree, perennial,* and *leaf.* It can be seen that the initial specification mark does not resolve the lexical ambiguity, since the word *plant* appears in two subhierarchies with different senses. The specification mark identified by {*plant#2, flora#2*}, however, contains the highest number of words (three) from the context and will therefore be the one chosen to resolve the sense two of the word *plant.* The words *tree* and *perennial* are also disambiguated, choosing for both the sense one. The word *leaf* does not appear in the subhierarchy of the specification mark {*plant#2, flora#2*}, and therefore this word has not been disambiguated. These words are beyond the scope of the disambiguation algorithm. They will be left aside to be processed by a complementary set of heuristics (see section 3.1.2).

### 3.1.1 DISAMBIGUATION ALGORITHM

In this section, we formally describe the SM algorithm which consists of the following five steps:

**Step 1:**
All nouns are extracted from a given context. These nouns constitute the input context, $Context = \{w_1, w_2, ..., w_n\}$. For example, $Context = \{plant, tree, perennial, leaf\}$.

**Step 2:**
For each noun $w_i$ in the context, all its possible senses $S_i = \{S_{i1}, S_{i2}, ..., S_{in}\}$ are obtained from WordNet. For each sense $S_{ij}$, the hypernym chain is obtained and stored in order into stacks. For example, Table 1 shows all the hypernyms synsets for each sense of the word *Plant.*

**Step 3:**
To each sense appearing in the stacks, the method associates the list of subsumed senses from the context (see Figure 2, which illustrates the list of subsumed senses for *plant#1* and *plant#2*).

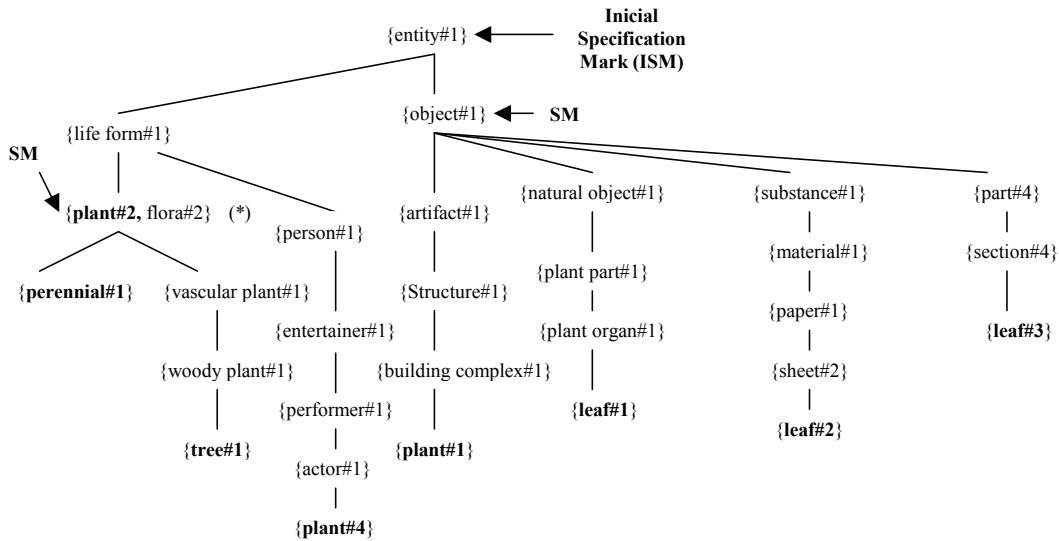

Figure 1: Specification Marks

| plant#1 | plant#2 | plant#3 | plant#4 |
|---|---|---|---|
| building complex#1 | life form#1 | contrivance#3 | actor#1 |
| structure#1 | entity#1 | scheme#1 | performer#1 |
| artifact#1 | | plan of action#1 | entertainer#1 |
| object#1 | | plan#1 | person#1 |
| entity#1 | | idea#1 | life form#1 |
| | | content#5 | entity#1 |
| | | cognition#1 | |
| | | psychological feature#1 | |

Table 1: Hypernyms synsets of *plant*

**Step 4:**

Beginning from the initial specification marks (the top synsets), the program descends recursively through the hierarchy, from one level to another, assigning to each specification mark the number of context words subsumed.

Figure 3 shows the word counts for *plant#1* through *plant#4* located within the specification mark *entity#1, ..., life form#1, flora#2*. For the *entity#1* specification mark, senses #1, #2, and #4 have the same maximal word counts (4). Therefore, it is not possible to disambiguate the word *plant* using the *entity#1* specification mark, and it will be necessary to go down one level of the hyponym hierarchy by changing the specification mark. Choosing the specification mark *life form#1*, senses #2 and #4 of *plant* have the same maximal word counts (3). Finally, it is possible to disambiguate the word *plant* with the sense #2 using the {plant#2, flora#2} specification mark, because of this sense has the higher word density (in this case, 3).

```
For PLANT:
    For PLANT#1:
        plant#1 → plant#1
        building complex#1 → plant#1
        structure#1 → plant#1
        artifact#1 → plant#1
        object#1 → plant#1, leaf#1, leaf#2, leaf#3
        entity#1 → plant#1, plant#2, plant#4, tree#1, perennial#1, leaf#1, leaf#2, leaf#3
    For PLANT#2:
        plant#2 → plant#2, tree#1, perennial#1
        life form#1 → plant#2, plant#4, tree#1, perennial#1
        entity#1 → plant#1, plant#2, plant#4, tree#1, perennial#1, leaf#1, leaf#2, leaf#3
```

Figure 2: Data Structure for Senses of the Word *Plant*

```
For PLANT
    located within the specification mark {entity#1}
        For PLANT#1 : 4 (plant, tree, perennial, leaf)
        For PLANT#2 : 4 (plant, tree, perennial, leaf)
        For PLANT#3 : 1 (plant)
        For PLANT#4 : 4 (plant, tree, perennial, leaf)
                ..............
    located within the specification mark {life form#1}
        For PLANT#1 : 1 (plant)
        For PLANT#2 : 3 (plant, tree, perennial)
        For PLANT#3 : 1 (plant)
        For PLANT#4 : 3 (plant, tree, perennial)
    located within the specification mark {plant #2, flora#2}
        For PLANT#1 : 1 (plant)
        For PLANT#2 : 3 (plant, tree, perennial)
        For PLANT#3 : 1 (plant)
        For PLANT#4 : 1 (plant)
```

Figure 3: Word Counts for Four Senses of the Word *Plant*

**Step 5:**
In this step, the method selects the word sense(s) having the greatest number of words counted in *Step 4*. If there is only one sense, then that is the one that is obviously chosen. If there is more than one sense, we repeat *Step 4*, moving down each level within the taxonomy until a single sense is obtained or the program reach a leaf specification mark. Figure 3 shows the word counts for each sense of *plant* (#1 through #4) located within the specification mark *entity#1, ..., life form#1, flora#2*. If the word cannot be disambiguated in this way, then it will be necessary to continue the disambiguation process applying a complementary set of heuristics.

### 3.1.2 Heuristics

The specification marks method is combined with a set of five knowledge-based heuristics: hypernym/hyponym, definition, gloss hypernym/hyponym, common specification mark, and domain heuristics. A short description of each of these methods is provided below.

### 3.1.3 Hypernym/Hyponym Heuristic

This heuristic solves the ambiguity of those words that are not explicitly related in WordNet (i.e., *leaf* is not directly related to *plant*, but rather follows a hypernym chain plus a PART-OF relation). All the hypernyms/hyponyms of the ambiguous word are checked, looking for synsets that have compounds that match with some word from the context. Each synset in the hypernym/hyponym chain is weighted in accordance with its depth within the subhierarchy. The sense then having the greatest weight is chosen. Figure 4 shows that, *leaf#1* being a hyponym of *plant_organ#1* is disambiguated (obtain the greatest weight, $weight(leaf\#1) = \sum_{i=1}^{depth} (\frac{level}{total\ levels}) = (\frac{4}{6}) + (\frac{5}{6}) = 1.5$) because *plant* is contained within the context of *leaf*.

**Context**: *plant, tree, leaf, perennial*
**Word non disambiguated**: *leaf.*
**Senses**: *leaf#1, leaf#2, leaf#3.*

For leaf#1

| | |
|---|---|
| => entity, something | *Level 1* |
| => object, physical object | *Level 2* |
| => natural object | *Level 3* |
| => **plant** part | *Level 4* |
| => **plant** organ | *Level 5* |
| => leaf#1, leafage, foliage | *Level 6* |

Figure 4: Application of the Hypernym Heuristic

### 3.1.4 Definition Heuristic

In this case, all the glosses from the synsets of an ambiguous word are checked looking for those that contain words from the context. Each match increases the synset count by one. The sense having the greatest count is then chosen. Figure 5 shows an example of this heuristic. The sense *sister#1* is chosen, because it has the greatest weight.

**Context**: *person, sister, musician.*
**Words non disambiguated**: *sister,musician.*
**Senses**: *sister#1, sister#2, sister#3 sister#4.*

For sister#1 → Weight = **2**

1. sister, sis -- (a female **person** who has the same parents as another **person**; "my sister married a **musician**")

For sister#3 → Weight = **1**

3. sister -- (a female **person** who is a fellow member (of a sorority or labor union or other group); "none of her sisters would betray her")

Figure 5: Application of the Definition Heuristic

### 3.1.5 Gloss Hypernym/Hyponym Heuristic

This method extends the previously defined hypernym/hyponym heuristic by using glosses of the hypernym/hyponym synsets of the ambiguous word. To disambiguate a given word, all the glosses of the hypernym/hyponym synsets are checked looking for words occurring in the context. Coincidences are counted. As before, the synset having the greatest count is chosen. Figure 6 shows a example of this heuristic. The sense *plane#1* is chosen, because it has the greatest weight.

**Context**: *plane, air*
**Words non disambiguated**: *plane*
**Senses**: *plane#1, plane#2, plane#3, plane#4, plane#5.*

For Plane#1: → Weight = **1**

airplane, aeroplane, plane -- (an aircraft that has fixed a wing and is powered by propellers or jets; "the flight was delayed due to trouble with the airplane")
    => aircraft -- (a vehicle that can fly)
      => craft -- (a vehicle designed for navigation in or on water or **<u>air</u>** or through outer space)
        => vehicle -- (a conveyance that transports people or objects)
          => conveyance, transport -- (something that serves as a means of transportation)
            => instrumentality, instrumentation -- (an artifact (or system of artifacts) that is instrumental in accomplishing some end)
            => artifact, artefact -- (a man-made object)
              => object, physical object -- (a physical (tangible and visible) entity; "it was full of rackets, balls and other objects")
              => entity, something -- (anything having existence (living or nonliving))

Figure 6: Application of the Gloss Hypernym Heuristic

### 3.1.6 Common Specification Mark Heuristic

In most cases, the senses of the words to be disambiguated are very close to each other and only differ in subtle differences in nuances. The Common Specification Mark heuristic reduce the ambiguity of a word without trying to provide a full disambiguation. Thus, we select the specification mark that is common to all senses of the context words, reporting all senses instead of choosing a single sense from among them. To illustrate this heuristic, consider Figure 7. In this example, the word *month* is not able to discriminate completely among four senses of the word *year*. However, in this case, the presence of the word *month* can help to select two possible senses of the word *year* when selecting the *time period, period* as a common specification mark. This specification mark represents the most specific common synset of a particular set of words. Therefore, this heuristic selects the sense *month#1* and senses *year#1* and *year#2* instead of attempting to choose a single sense or leaving them completely ambiguous.

### 3.1.7 Domain WSD Heuristic

This heuristic uses a derived resource, "relevant domains" (Montoyo et al., 2003), which is obtained combining both the WordNet glosses and WordNet Domains (Magnini & Strap-

**Context**: *year, month*. **Words non disambiguated**: *year*. **Senses**: *year#1, year#2, year#3, year#4*.

For year#1:                        For year#2:                        For month#1:

  => abstraction                  => abstraction                  => abstraction
    => measure, quantity            => measure, quantity            => measure, quantity
    **=> time period, period**        **=> time period, period**        **=> time period, period**
      => year#1, twelvemonth            => year#2                    => month#1

Figure 7: Example of Common Specification Mark Heuristic

parava, 2000)[3]. WordNet Domains establish a semantic relation between word senses by grouping them into the same semantic domain (*Sports, Medicine,* etc.). The word *bank*, for example, has ten senses in WordNet 2.0, but three of them, "bank#1", "bank#3" and "bank#6" are grouped into the same domain label, *Economy*, whereas "bank#2" and "bank#7" are grouped into the domain labels *Geography* and *Geology*. These domain labels are selected from a set of 165 labels hierarchically organized. In that way, a domain connects words that belong to different subhierarchies and part–of–speech.

"Relevant domains" is a lexicon derived from the WordNet glosses using WordNet Domains. In fact, we use WordNet as a corpus categorized with domain labels. For each English word appearing in the gloses of WordNet, we obtain a list of their most representative domain labels. The relevance is obtained weighting each possible label with the "Association Ratio" formula (AR), where $w$ is a word and $D$ is a domain.

$$AR(w|D) = P(w|D) * \log \frac{P(w|D)}{P(w)} \qquad (1)$$

This list can also be considered as a weighted vector (or point in a multidimensional space). Using such word vectors of "Relevant domains", we can derive new vectors to represent sets of words—for instance, for contexts or glosses. We can then compare the similarity between a given context and each of the possible senses of a polysemous word—by using for instance the cosine function.

Figure 8 shows an example for disambiguating the word *genotype* in the following text: *There are a number of ways in which the chromosome structure can change, which will detrimentally change the genotype and phenotype of the organism.* First, the glosses of the word to be disambiguated and the context are pos–tagged and analyzed morphologically. Second, we build the context vector (CV) which combines in one structure the most relevant and representative domains related to the words from the text to be disambiguated. Third, in the same way, we build the sense vectors (SV) which group the most relevant and representative domains of the gloss that is associated with each one of the word senses. In this example, *genotype#1 – (a group of organisms sharing a specific genetic constitution)* and *genotype#2 – (the particular alleles at specified loci present in an organism)*. Finally, in order to select the appropriate sense, we made a comparison between all sense vectors and the context vector, and we select the senses more approximate to the context vector. In this example, we show the sense vector for sense *genotype#1* and we select the *genotype#1* sense, because its cosine is higher.

---

3. http://wndomains.itc.it/

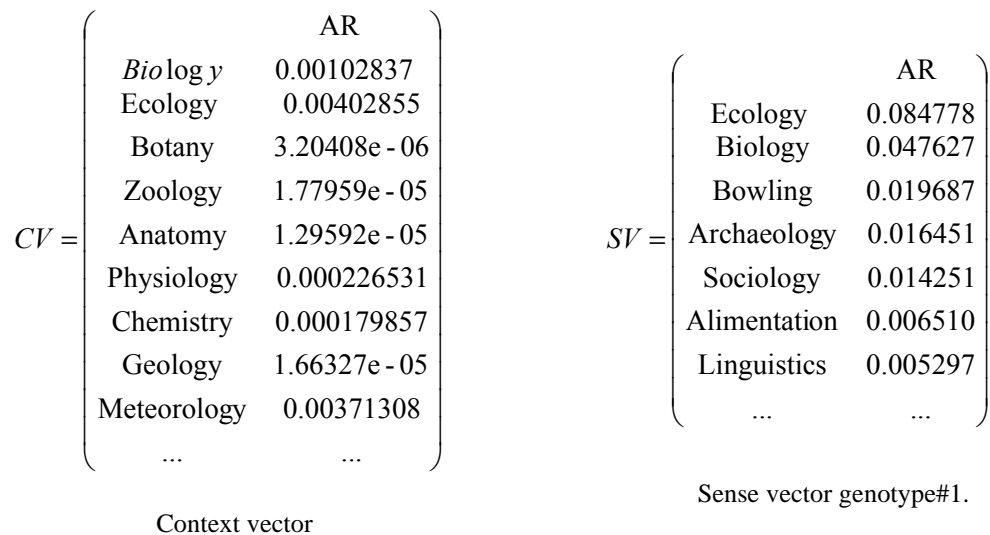

Context vector

Sense vector genotype#1.

Selected sense

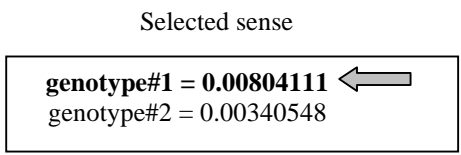

Figure 8: Example of Domain WSD Heuristic

Defining this heuristic as "knowledge-based" or "corpus-based" can be seen controversial because this heuristic uses WordNet gloses (and WordNet Domains) as a corpus to derive the "relevant domains". That is, using corpus techniques on WordNet. However, WordNet Domains was constructed semi-automatically (prescribed) following the hierarchy of WordNet.

### 3.1.8 Evaluation of Specification Marks Method

Obviously, we can also use different strategies to combine a set of knowledge-based heuristics. For instance, all the heuristics described in the previous section can be applied in order passing to the next heuristic only the remaining ambiguity that previous heuristics were not able to solve.

In order to evaluate the performance of the knowledge-based heuristics previously defined, we used the SemCor collection (Miller et al., 1993), in which all content words are annotated with the most appropriate WordNet sense.In this case, we used a window of fifteen nouns (seven context nouns before and after the target noun).

The results obtained for the specification marks method using the heuristics when applied one by one are shown in Table 2. This table shows the results for polysemous nouns only, and for polysemous and monosemous nouns combined.

| Heuristics | Precision | Recall | Coverage |
|---|---|---|---|
| Polysemic and monosemic nouns | 0.553 | 0.522 | 0.943 |
| Only polysemic nouns | 0.377 | 0.311 | 0.943 |

Table 2: Results Using Heuristics Applied in Order on SemCor

The results obtained for the heuristics applied independently are shown in Table 3. As shown, all the heuristics perform differently, providing different precision/recall figures.

| Heuristics | Precision | | Recall | | Coverage | |
|---|---|---|---|---|---|---|
| | Mono+Poly | Polysemic | Mono+Poly | Polysemic | Mono+Poly | Polysemic |
| Spec. Mark Method | 0.383 | 0.300 | 0.341 | 0.292 | 0.975 | 0.948 |
| Hypernym | 0.563 | 0.420 | 0.447 | 0.313 | 0.795 | 0.745 |
| Definition | 0.480 | 0.300 | 0.363 | 0.209 | 0.758 | 0.699 |
| Hyponym | 0.556 | 0.393 | 0.436 | 0.285 | 0.784 | 0.726 |
| Gloss hypernym | 0.555 | 0.412 | 0.450 | 0.316 | 0.811 | 0.764 |
| Gloss hyponym | 0.617 | 0.481 | 0.494 | 0.358 | 0.798 | 0.745 |
| Common specification | 0.565 | 0.423 | 0.443 | 0.310 | 0.784 | 0.732 |
| Domain WSD | 0.585 | 0.453 | 0.483 | 0.330 | 0.894 | 0.832 |

Table 3: Results Using Heuristics Applied Independently

Another possibility is to combine all the heuristics using a majority voting schema (Rigau et al., 1997). In this simple schema, each heuristic provides a vote, and the method selects the synset that obtains more votes. The results shown in Table 4 illustrate that when the heuristics are working independently, the method achieves a 39.1% recall for polysemous nouns (with full coverage), which represents an improvement of 8 percentual points over the method in which heuristics are applied in order (one by one).

| | Precision | | Recall | |
|---|---|---|---|---|
| | Mono+Poly | Polysemic | Mono+Poly | Polysemic |
| Voting heuristics | 0.567 | 0.436 | 0.546 | 0.391 |

Table 4: Results using majority voting on SemCor

We also show in Table 5 the results of our domain heuristic when applied on the English all-words task from Senseval-2. In the table, the polysemy reduction caused by domain clustering can profitably help WSD. Since domains are coarser than synsets, word domain disambiguation (WDD) (Magnini & Strapparava, 2000) can obtain better results than WSD. Our goal is to perform a preliminary domain disambiguation in order to provide an informed search–space reduction.

### 3.1.9 Comparison with Knowledge-based Methods

In this section we compare three different knowledge-based methods: conceptual density (Agirre & Rigau, 1996), a variant of the conceptual density algorithm (Fernández-Amorós et al., 2001); the Lesk method (Lesk, 1986) ; and the specification marks method.

| Level WSD | Precision | Recall |
|:---:|:---:|:---:|
| Sense | 0.44 | 0.32 |
| Domain | 0.54 | 0.43 |

Table 5: Results of Use of Domain WSD Heuristic

Table 6 shows recall results for the three methods when applied to the entire SemCor collection. Our best result achieved 39.1% recall. This is an important improvement with respect to other methods, but the results are still far below the most frequent sense heuristic. Obviously, none of the knowledge-based techniques and heuristics presented above are sufficient, in isolation, to perform accurate WSD. However, we have empirically demonstrated that a simple combination of knowledge-based heuristics can lead to improvements in the WSD process.

| WSD Method | Recall |
|:---|:---:|
| SM and Voting Heuristics | 0.391 |
| UNED Method | 0.313 |
| SM with Cascade Heuristics | 0.311 |
| Lesk | 0.274 |
| Conceptual Density | 0.220 |

Table 6: Recall results using three different knowledge–based WSD methods

## 3.2 Maximum Entropy-based Method

Maximum Entropy modeling provides a framework for integrating information for classification from many heterogeneous information sources (Manning & Schütze, 1999; Berger et al., 1996). ME probability models have been successfully applied to some NLP tasks, such as POS tagging or sentence-boundary detection (Ratnaparkhi, 1998).

The WSD method used in this work is based on conditional ME models. It has been implemented using a supervised learning method that consists of building word-sense classifiers using a semantically annotated corpus. A classifier obtained by means of an ME technique consists of a set of parameters or coefficients which are estimated using an optimization procedure. Each coefficient is associated with one feature observed in the training data. The goal is to obtain the probability distribution that maximizes the entropy—that is, maximum ignorance is assumed and nothing apart from the training data is considered. One advantage of using the ME framework is that even knowledge-poor features may be applied accurately; the ME framework thus allows a virtually unrestricted ability to represent problem-specific knowledge in the form of features (Ratnaparkhi, 1998).

Let us assume a set of contexts $X$ and a set of classes $C$. The function $cl : X \rightarrow C$ chooses the class $c$ with the highest conditional probability in the context $x$: $cl(x) = \arg\max_c p(c|x)$. Each feature is calculated by a function that is associated with a specific class $c'$, and it takes the form of equation (2), where $cp(x)$ represents some observable characteristic in the context[4]. The conditional probability $p(c|x)$ is defined by equation (3), where $\alpha_i$ is the parameter or weight of the feature $i$, $K$ is the number of features defined, and $Z(x)$ is a normalization factor that ensures that the sum of all conditional probabilities for this context is equal to 1.

$$f(x,c) = \begin{cases} 1 & \text{if } c' = c \text{ and } cp(x) = true \\ 0 & \text{otherwise} \end{cases} \quad (2)$$

$$p(c|x) = \frac{1}{Z(x)} \prod_{i=1}^{K} \alpha_i^{f_i(x,c)} \quad (3)$$

The learning module produces the classifiers for each word using a corpus that is syntactically and semantically annotated. The module processes the learning corpus in order to define the functions that will apprise the linguistic features of each context.

For example, consider that we want to build a classifier for the noun *interest* using the POS label of the previous word as a feature and we also have the the following three examples from the training corpus:

... the widespread **interest#1** in the ...
... the best **interest#5** of both ...
... persons expressing **interest#1** in the ...

The learning module performs a sequential processing of this corpus, looking for the pairs *<POS-label, sense>*. Then, the following pairs are used to define three functions (each context has a vector composed of three features).

*<adjective,#1>*
*<adjective,#5>*
*<verb,#1>*

We can define another type of feature by merging the POS occurrences by sense:

*< {adjective,verb},#1>*
*<adjective,#5>*

This form of defining the pairs means a reduction of feature space because all information (of some kind of linguistic data, e.g., POS label at position -1) about a sense is contained in just one feature. Obviously, the form of the feature function 2 must be adapted to Equation 4. Thus,

$$W_{(c')} = \{data \ of \ sense \ c'\} \quad (4)$$

$$f_{(c',i)}(x,c) = \begin{cases} 1 & \text{if } c' = c \text{ and } CP(x) \in W_{(c')} \\ 0 & \text{otherwise} \end{cases}$$

---

4. The ME approach is not limited to binary features, but the optimization procedure used for the estimation of the parameters, the *Generalized Iterative Scaling* procedure, uses this kind of features.

We will refer to the feature function expressed by Equation 4 as "collapsed features". The previous Equation 2 we call "non-collapsed features". These two feature definitions are complementary and can be used together in the learning phase.

Due to the nature of the disambiguation task, the number of times that a feature generated by the first type of function ("non-collapsed") is activated is very low, and the feature vectors have a large number of null values. The new function drastically reduces the number of features, with a minimal degradation in the evaluation results. In this way, more and new features can be incorporated into the learning process, compensating the loss of accuracy.

Therefore, the classification module carries out the disambiguation of new contexts using the previously stored classification functions. When ME does not have enough information about a specific context, several senses may achieve the same maximum probability and thus the classification cannot be done properly. In these cases, the most frequent sense in the corpus is assigned. However, this heuristic is only necessary for a minimum number of contexts or when the set of linguistic attributes processed is very small.

### 3.2.1 DESCRIPTION OF FEATURES

The set of features defined for the training of the system is described in Figure 9 and is based on the features described by Ng and Lee (1996) and Escudero et al. (2000). These features represent words, collocations, and POS tags in the local context. Both "collapsed" and "non-collapsed" functions are used.

- $\mathbf{0}$: word form of the target word
- $\mathbf{s}$: words at positions $\pm 1$, $\pm 2$, $\pm 3$
- $\mathbf{p}$: POS-tags of words at positions $\pm 1$, $\pm 2$, $\pm 3$
- $\mathbf{b}$: lemmas of collocations at positions $(-2, -1)$, $(-1, +1)$, $(+1, +2)$
- $\mathbf{c}$: collocations at positions $(-2, -1)$, $(-1, +1)$, $(+1, +2)$
- $\mathbf{k}$m: lemmas of nouns at any position in context, occurring at least $m\%$ times with a sense
- $\mathbf{r}$: grammatical relation to the ambiguous word
- $\mathbf{d}$: the word that the ambiguous word depends on
- $\mathbf{m}$: multi-word if identified by the parser
- $\mathbf{L}$: lemmas of content-words at positions $\pm 1$, $\pm 2$, $\pm 3$ ("collapsed" definition)
- $\mathbf{W}$: content-words at positions $\pm 1$, $\pm 2$, $\pm 3$ ("collapsed" definition)
- $\mathbf{S}$, $\mathbf{B}$, $\mathbf{C}$, $\mathbf{P}$, $\mathbf{D}$ and $\mathbf{M}$: "collapsed" versions (see Equation 4)

Figure 9: Features Used for the Training of the System

Actually, each item in Figure 9 groups several sets of features. The majority of them depend on the nearest words (e.g., $s$ comprises all possible features defined by the words occurring in each sample at positions $w_{-3}$, $w_{-2}$, $w_{-1}$, $w_{+1}$, $w_{+2}$, $w_{+3}$ related to the ambiguous word). Types nominated with capital letters are based on the "collapsed" function form; that is, these features simply recognize an attribute belonging to the training data.

Keyword features ($k$m) are inspired by Ng and Lee work. Noun filtering is done using frequency information for nouns co-occurring with a particular sense. For example, let us

suppose $m = 10$ for a set of 100 examples of *interest#4*: if the noun *bank* is found 10 times or more at any position, then a feature is defined.

Moreover, new features have also been defined using other grammatical properties: relationship features ($r$) that refer to the grammatical relationship of the ambiguous word (*subject*, *object*, *complement*, ...) and dependency features ($d$ and $D$) that extract the word related to the ambiguous one through the dependency parse tree.

### 3.2.2 Evaluation of the Maximum Entropy Method

In this subsection we present the results of our evaluation over the training and testing data of the Senseval-2 Spanish lexical–sample task. This corpus was parsed using *Conexor Functional Dependency Grammar parser for Spanish* (Tapanainen & Järvinen, 1997).

The classifiers were built from the training data and evaluated over the test data. Table 7 shows which combination of groups of features works better for every POS and which work better for all words together.

|  | Accuracy | Feature selection |
|---|---|---|
| Nouns | 0.683 | LWSBCk5 |
| Verbs | 0.595 | sk5 |
| Adjectives | 0.783 | LWsBCp |
| ALL | 0.671 | 0LWSBCk5 |

Table 7: Baseline: Accuracy Results of Applying ME on Senseval-2 Spanish Data

This work entails an exhaustive search looking for the most accurate combination of features. The values presented here are merely informative and indicate the maximum accuracy that the system can achieve with a particular set of features.

## 3.3 Improving ME accuracy

Our main goal is to find a method that will automatically obtain the best feature selection (Veenstra et al., 2000; Mihalcea, 2002; Suárez & Palomar, 2002b) from the training data. We performed an 3-fold cross-validation process. Data is divided in 3 folds; then, 3 tests are done, each one with 2 folds as training data and the remaining one as testing data. The final result is the average accuracy. We decided on just three tests because of the small size of the training data. Then, we tested several combinations of features over the training data of the Senseval-2 Spanish lexical–sample and analyzed the results obtained for each word.

In order to perform the 3-fold cross-validation process on each word, some preprocessing of the corpus was done. For each word, all senses were uniformly distributed into the three folds (each fold contains one-third of the examples of each sense). Those senses that had fewer than three examples in the original corpus file were rejected and not processed.

Table 8 shows the best results obtained using three-fold cross-validation on the training data. Several feature combinations were tested in order to find the best set for each selected word. The purpose was to obtain the most relevant information for each word from the corpus rather than applying the same combination of features to all of them. Therefore, the information in the column *Features* lists only the feature selection with the best result.

| Word | Features | Accur | MFS | Word | Features | Accur | MFS |
|---|---|---|---|---|---|---|---|
| autoridad,N | sbcp | 0.589 | 0.503 | clavar,V | sbcprdk3 | 0.561 | 0.449 |
| bomba,N | 0LWSBCk5 | 0.762 | 0.707 | conducir,V | LWsBCPD | 0.534 | 0.358 |
| canal,N | sbcprdk3 | 0.579 | 0.307 | copiar,V | 0sbcprdk3 | 0.457 | 0.338 |
| circuito,N | 0LWSBCk5 | 0.536 | 0.392 | coronar,V | sk5 | 0.698 | 0.327 |
| corazón,N | 0Sbcpk5 | 0.781 | 0.607 | explotar,V | 0LWSBCk5 | 0.593 | 0.318 |
| corona,N | sbcp | 0.722 | 0.489 | saltar,V | LWsBC | 0.403 | 0.132 |
| gracia,N | 0sk5 | 0.634 | 0.295 | tocar,V | 0sbcprdk3 | 0.583 | 0.313 |
| grano,N | 0LWSBCr | 0.681 | 0.483 | tratar,V | sbcpk5 | 0.527 | 0.208 |
| hermano,N | 0Sprd | 0.731 | 0.602 | usar,V | 0Sprd | 0.732 | 0.669 |
| masa,N | LWSBCk5 | 0.756 | 0.455 | vencer,V | sbcprdk3 | 0.696 | 0.618 |
| naturaleza,N | sbcprdk3 | 0.527 | 0.424 | brillante,A | sbcprdk3 | 0.756 | 0.512 |
| operación,N | 0LWSBCk5 | 0.543 | 0.377 | ciego,A | 0spdk5 | 0.812 | 0.565 |
| órgano,N | 0LWSBCPDk5 | 0.715 | 0.515 | claro,A | 0Sprd | 0.919 | 0.854 |
| partido,N | 0LWSBCk5 | 0.839 | 0.524 | local,A | 0LWSBCr | 0.798 | 0.750 |
| pasaje,N | sk5 | 0.685 | 0.451 | natural,A | sbcprdk10 | 0.471 | 0.267 |
| programa,N | 0LWSBCr | 0.587 | 0.486 | popular,A | sbcprdk10 | 0.865 | 0.632 |
| tabla,N | sk5 | 0.663 | 0.488 | simple,A | LWsBCPD | 0.776 | 0.621 |
| actuar,V | sk5 | 0.514 | 0.293 | verde,A | LWSBCk5 | 0.601 | 0.317 |
| apoyar,V | 0sbcprdk3 | 0.730 | 0.635 | vital,A | Sbcp | 0.774 | 0.441 |
| apuntar,V | 0LWsBCPDk5 | 0.661 | 0.478 | | | | |

Table 8: Three-fold Cross-Validation Results on Senseval-2 Spanish Training Data: Best Averaged Accuracies per Word

Strings in each row represent the entire set of features used when training each classifier. For example, *autoridad* obtains its best result using nearest words, collocations of two lemmas, collocations of two words, and POS information that is, $s$, $b$, $c$, and $p$ features, respectively (see Figure 9). The column *Accur* (for "accuracy") shows the number of correctly classified contexts divided by the total number of contexts (because ME always classifies precision as equal to recall). Column *MFS* shows the accuracy obtained when the most frequent sense is selected.

The data summarized in Table 8 reveal that using "collapsed" features in the ME method is useful; both "collapsed" and "non-collapsed" functions are used, even for the same word. For example, the adjective *vital* obtains the best result with "*Sbcp*" (the "collapsed" version of words in a window $(-3.. + 3)$, collocations of two lemmas and two words in a window $(-2.. + 2)$, and POS labels, in a window $(-3.. + 3)$ too); we can here infer that single-word information is less important than collocations in order to disambiguate *vital* correctly.

The target word (feature 0) is useful for nouns, verbs, and adjectives, but many of the words do not use it for their best feature selection. In general, these words do not have a relevant relationship between shape and senses. On the other hand, POS information ($p$ and $P$ features) is selected less often. When comparing *lemma* features with *word* features (e.g., $L$ versus $W$, and $B$ versus $C$), they are complementary in the majority of cases. Grammatical relationships ($r$ features) and word–word dependencies ($d$ and $D$ features) seem very useful, too, if combined with other types of attributes. Moreover, keywords ($km$ features) are used very often, possibly due to the source and size of contexts of Senseval-2 Spanish lexical–sample data.

Table 9 shows the best feature selections for each part-of-speech and for all words. The data presented in Tables 8 and 9 were used to build four different sets of classifiers in order to compare their accuracy: **MEfix** uses the overall best feature selection for all words; **MEbfs** trains each word with its best selection of features (in Table 8); **MEbfs.pos** uses the best selection per POS for all nouns, verbs and adjectives, respectively (in Table 9); and, finally, **vME** is a majority voting system that has as input the answers of the preceding systems.

| POS | Acc | Features | System |
|---|---|---|---|
| Nouns | 0.620 | LWSBCk5 | |
| Verbs | 0.559 | sbcprdk3 | *MEbfs.pos* |
| Adjectives | 0.726 | 0spdk5 | |
| ALL | 0.615 | sbcprdk3 | *MEfix* |

Table 9: Three-fold Cross-Validation Results on Senseval-2 Spanish Training Data: Best Averaged Accuracies per POS

Table 10 shows a comparison of the four systems. **MEfix** has the lower results. This classifier applies the same set of types of features to all words. However, the best feature selection per word (**MEbfs**) is not the best, probably because more training examples are necessary. The best choice seems to select a fixed set of types of features for each POS (**MEbfs.pos**).

| ALL | | Nouns | |
|---|---|---|---|
| 0.677 | MEbfs.pos | 0.683 | MEbfs.pos |
| 0.676 | vME | 0.678 | vME |
| 0.667 | MEbfs | 0.661 | MEbfs |
| 0.658 | MEfix | 0.646 | MEfix |
| Verbs | | Adjectives | |
| 0.583 | vME | 0.774 | vME |
| 0.583 | MEbfs.pos | 0.772 | MEbfs.pos |
| 0.583 | MEfix | 0.771 | MEbfs |
| 0.580 | MEbfs | 0.756 | MEfix |

**MEfix**: *sbcprdk3* for all words
**MEbfs**: each word with its best feature selection
**MEbfs.pos**: *LWSBCk5* for nouns, *sbcprdk3* for verbs, and *0spdk5* for adjectives
**vME**: majority voting between MEfix, MEbfs.pos, and MEbfs

Table 10: Evaluation of ME Systems

While **MEbfs** predicts, for each word over the training data, which individually selected features could be the best ones when evaluated on the testing data, **MEbfs.pos** is an averaged prediction, a selection of features that, over the training data, performed a "good enough" disambiguation of the majority of words belonging to a particular POS. When this averaged prediction is applied to the real testing data, **MEbfs.pos** performs better than **MEbfs**.

Another important issue is that **MEbfs.pos** obtains an accuracy slightly better than the best possible evaluation result achieved with ME (see Table 7)—that is, a *best-feature-selection per POS* strategy from training data guarantees an improvement on ME-based WSD.

In general, verbs are difficult to learn and the accuracy of the method for them is lower than for other POS; in our opinion, more information (knowledge-based, perhaps) is needed to build their classifiers. In this case, the voting system (**vME**) based on the agreement between the other three systems, does not improve accuracy.

Finally in Table 11, the results of the ME method are compared with those systems that competed at Senseval-2 in the Spanish lexical–sample task[5]. The results obtained by ME systems are excellent for nouns and adjectives, but not for verbs. However, when comparing ALL POS, the ME systems seem to perform comparable to the best Senseval-2 systems.

| | ALL | | Nouns | | Verbs | | Adjectives |
|---|---|---|---|---|---|---|---|
| 0.713 | jhu(R) | 0.702 | jhu(R) | 0.643 | jhu(R) | 0.802 | jhu(R) |
| 0.682 | jhu | 0.683 | **MEbfs.pos** | 0.609 | jhu | 0.774 | **vME** |
| 0.677 | **MEbfs.pos** | 0.681 | jhu | 0.595 | css244 | 0.772 | **MEbfs.pos** |
| 0.676 | **vME** | 0.678 | **vME** | 0.584 | umd-sst | 0.772 | css244 |
| 0.670 | css244 | 0.661 | **MEbfs** | 0.583 | **vME** | 0.771 | **MEbfs** |
| 0.667 | **MEbfs** | 0.652 | css244 | 0.583 | **MEbfs.pos** | 0.764 | jhu |
| 0.658 | **MEfix** | 0.646 | **MEfix** | 0.583 | **MEfix** | 0.756 | **MEfix** |
| 0.627 | umd-sst | 0.621 | duluth 8 | 0.580 | **MEbfs** | 0.725 | duluth 8 |
| 0.617 | duluth 8 | 0.612 | duluth Z | 0.515 | duluth 10 | 0.712 | duluth 10 |
| 0.610 | duluth 10 | 0.611 | duluth 10 | 0.513 | duluth 8 | 0.706 | duluth 7 |
| 0.595 | duluth Z | 0.603 | umd-sst | 0.511 | ua | 0.703 | umd-sst |
| 0.595 | duluth 7 | 0.592 | duluth 6 | 0.498 | duluth 7 | 0.689 | duluth 6 |
| 0.582 | duluth 6 | 0.590 | duluth 7 | 0.490 | duluth Z | 0.689 | duluth Z |
| 0.578 | duluth X | 0.586 | duluth X | 0.478 | duluth X | 0.687 | ua |
| 0.560 | duluth 9 | 0.557 | duluth 9 | 0.477 | duluth 9 | 0.678 | duluth X |
| 0.548 | ua | 0.514 | duluth Y | 0.474 | duluth 6 | 0.655 | duluth 9 |
| 0.524 | duluth Y | 0.464 | ua | 0.431 | duluth Y | 0.637 | duluth Y |

Table 11: Comparison with the Spanish Senseval-2 systems

---

5. JHU(R) by Johns Hopkins University; CSS244 by Stanford University; UMD-SST by the University of Maryland; Duluth systems by the University of Minnesota - Duluth; UA by the University of Alicante.

### 3.4 Comparing Specification Marks Method with Maximum Entropy-based Method

The main goal of this section is to evaluate the Specification Marks Method (Montoyo & Suárez, 2001) and the Maximum Entropy-based Method (in particular, **MEfix** System) on a common data set, to allow for direct comparisons. The individual evaluation of each method has been carried out on the noun set (17 nouns) of the Spanish lexical-sample task (Rigau et al., 2001) from Senseval-2[6]. Table 12 shows precision, recall and coverage of both methods.

|    | Precision | Recall | Coverage |
|----|-----------|--------|----------|
| SM | 0.464     | 0.464  | 0.941    |
| ME | 0.646     | 0.646  | 1        |

Table 12: Comparison of ME and SM for nouns in Senseval-2 Spanish lexical sample

In order to study a possible cooperation between both methods, we count those cases that: the two methods return the correct sense for the same occurrence, at least one of the methods provides the correct sense and finally, none of both provides the correct sense. A summary of the obtained results is shown in the Table 13. These results clearly show that there is a large room for improvement when combining both system outcomes. In fact, they provide also a possible upper bound precision for this technology, which can be set to 0.798 (more than 15 percentual points higher than the current best system). Table 14 presents a complementary view: wins, ties and loses between ME and SM when each context is examined. Although ME performs better than SM, there are 122 cases (15 %) solved only by the SM method.

|         | Contexts | Percentage |
|---------|----------|------------|
| Both OK | 240      | 0.300      |
| One OK  | 398      | 0.498      |
| Zero OK | 161      | 0.202      |

Table 13: Correct classifications of ME and SM for nouns in Senseval-2 Spanish lexical sample

|         | Wins | Ties | Loses |
|---------|------|------|-------|
| ME – SM | 267  | 240  | 122   |

Table 14: Wins, ties and loses of ME and SM systems for nouns in Senseval-2 Spanish lexical sample

---

6. For both Spanish and English Senseval-2 corpora, when applying the Specification Marks method we used the whole example as the context window for the target noun

## 4. Experimental Work

In this section we attempt to confirm our hypothesis that both corpus-based and knowledge-based methods can improve the accuracy of each other. The first subsection shows the results of preprocessing the test data with the maximum entropy method (ME) in order to help the specification marks method (SM). Next, we test the opposite, if preprocessing the test data with the domain heuristic can help the maximum entropy method to disambiguate more accurately.

The last experiment combines the vME system (the majority voting system) and SM method. Actually, it relies on simply adding the SM as one more heuristic to the voting scheme.

### 4.1 Integrating a Corpus-based WSD System into a Knowledge-based WSD System

This experiment was designed to study and evaluate whether the integration of corpus-based system within a knowledge-based helps improve word-sense disambiguation of nouns.

Therefore, ME can help to SM by labelling some nouns of the context of the target word. That is, reducing the number of possible senses of some nouns of the context. In fact, we reduce the search space of the SM method. This ensures that the sense of the target word will be the one more related to the noun senses labelled by ME.

In this case, we used the noun words from the English lexical-sample task from SENSEVAL-2. ME helps SM by labelling some words from the context of the target word. These words were sense tagged using the SemCor collection as a learning corpus. We performed a three–fold cross-validation for all nouns having 10 or more occurrences. We selected those nouns that were disambiguated by ME with high precision, that is, nouns that had percentage rates of accuracy of 90% or more. The classifiers for these nouns were used to disambiguate the testing data. The total number of different noun classifiers (noun) activated for each target word across the testing corpus is shown in Table 15.

Next, SM was applied, using all the heuristics for disambiguating the target words of the testing data, but with the advantage of knowing the senses of some nouns that formed the context of these targets words.

Table 15 shows the results of precision and recall when SM is applied with and without first applying ME, that is, with and without fixing the sense of the nouns that form the context. A very small but consistent improvement was obtained through the complete test set (3.56% precision and 3.61% recall). Although the improvement is very low, this experiment empirically demonstrates that a corpus-based method such as maximum entropy can be integrated to help a knowledge-based system such as the specification marks method.

### 4.2 Integrating a Knowledge-based WSD system into a Corpus-based WSD system

In this case, we used only the domain heuristic to improve ME because this information can be added directly as domain features. The problem of data sparseness from which a WSD system based on features suffers could be increased by the fine-grained sense distinctions provided by WordNet. On the contrary, the domain information significantly reduces the word polysemy (i.e., the number of categories for a word is generally lower than the number of senses for the word) and the results obtained by this heuristic have better precision than those obtained by the whole SM method, which in turn obtain better recall.

| Target words | noun classifiers | Without fixed senses | | With fixed senses | |
|---|---|---|---|---|---|
| | | Precision | Recall | Precision | Recall |
| art | 63 | 0.475 | 0.475 | 0.524 | 0.524 |
| authority | 80 | 0.137 | 0.123 | 0.144 | 0.135 |
| bar | 104 | 0.222 | 0.203 | 0.232 | 0.220 |
| bum | 37 | 0.421 | 0.216 | 0.421 | 0.216 |
| chair | 59 | 0.206 | 0.190 | 0.316 | 0.301 |
| channel | 32 | 0.500 | 0.343 | 0.521 | 0.375 |
| child | 59 | 0.500 | 0.200 | 0.518 | 0.233 |
| church | 50 | 0.509 | 0.509 | 0.540 | 0.529 |
| circuit | 49 | 0.356 | 0.346 | 0.369 | 0.360 |
| day | 136 | 0.038 | 0.035 | 0.054 | 0.049 |
| detention | 22 | 0.454 | 0.454 | 0.476 | 0.454 |
| dyke | 15 | 0.933 | 0.933 | 0.933 | 0.933 |
| facility | 14 | 0.875 | 0.875 | 1 | 1 |
| fatigue | 38 | 0.236 | 0.230 | 0.297 | 0.282 |
| feeling | 48 | 0.306 | 0.300 | 0.346 | 0.340 |
| grip | 38 | 0.184 | 0.179 | 0.216 | 0.205 |
| hearth | 29 | 0.321 | 0.310 | 0.321 | 0.310 |
| holiday | 23 | 0.818 | 0.346 | 0.833 | 0.384 |
| lady | 40 | 0.375 | 0.136 | 0.615 | 0.363 |
| material | 58 | 0.343 | 0.338 | 0.359 | 0.353 |
| mouth | 51 | 0.094 | 0.094 | 0.132 | 0.132 |
| nation | 25 | 0.269 | 0.269 | 0.307 | 0.307 |
| nature | 37 | 0.263 | 0.263 | 0.289 | 0.289 |
| post | 41 | 0.312 | 0.306 | 0.354 | 0.346 |
| restraint | 31 | 0.200 | 0.193 | 0.206 | 0.193 |
| sense | 37 | 0.260 | 0.240 | 0.282 | 0.260 |
| spade | 17 | 0.823 | 0.823 | 0.941 | 0.941 |
| stress | 37 | 0.228 | 0.216 | 0.257 | 0.243 |
| yew | 24 | 0.480 | 0.480 | 0.541 | 0.520 |
| Total | 1294 | 0.300 | 0.267 | 0.336 | 0.303 |

Table 15: Precision and Recall Results Using SM to Disambiguate Words, With and Without Fixing of Noun Sense

As shown in subsection 3.1.7, the domain heuristic can annotate word senses characterized by their domains. Thus, these domains will be used as an additional type of features for ME in a context window of $\pm 1$, $\pm 2$, and $\pm 3$ from the target word. In addition, the three more relevant domains were calculated also for each context and incorporated to the training in the form of features.

This experiment was also carried out on the English lexical-sample task data from Senseval-2, and ME was used to generate two groups of classifiers from the training data.

The first group of classifiers used the corpus without information of domains; the second, having previously been domain disambiguated with SM, incorporating the domain label of adjacent nouns, and the three more relevant domains to the context. That is, providing to the classifier a richer set of features (adding the domain features). However, in this case, we did not perform any feature selection.

The test data was disambiguated by ME twice, with and without SM domain labelling, using $0lWsbcpdm$ (see Figure 9) as the common set of features in order to perform the comparison. The results of the experiment are shown in Table 16.

The table shows that 7 out of 29 nouns obtained worse results when using the domains, whereas 13 obtained better results. Although, in this case, we only obtained a very small improvement in terms of precision (2%)[7].

---

7. This difference proves to be statistically significant when applying the test of the corrected difference of two proportions (Dietterich, 1998; Snedecor & Cochran, 1989)

| Target words | Without domains | With domains | Improvement |
|---|---|---|---|
| art | 0.667 | 0.778 | 0.111 |
| authority | 0.600 | 0.700 | 0.100 |
| bar | 0.625 | 0.615 | -0.010 |
| bum | 0.865 | 0.919 | 0.054 |
| chair | 0.898 | 0.898 | |
| channel | 0.567 | 0.597 | 0.030 |
| child | 0.661 | 0.695 | 0.034 |
| church | 0.560 | 0.600 | 0.040 |
| circuit | 0.408 | 0.388 | -0.020 |
| day | 0.676 | 0.669 | -0.007 |
| detention | 0.909 | 0.909 | |
| dyke | 0.800 | 0.800 | |
| facility | 0.429 | 0.500 | 0.071 |
| fatigue | 0.850 | 0.850 | |
| feeling | 0.708 | 0.688 | -0.021 |
| grip | 0.540 | 0.620 | 0.080 |
| hearth | 0.759 | 0.793 | 0.034 |
| holiday | 1.000 | 0.957 | -0.043 |
| lady | 0.900 | 0.900 | |
| material | 0.534 | 0.552 | 0.017 |
| mouth | 0.569 | 0.588 | 0.020 |
| nation | 0.720 | 0.720 | |
| nature | 0.459 | 0.459 | |
| post | 0.463 | 0.512 | 0.049 |
| restraint | 0.516 | 0.452 | -0.065 |
| sense | 0.676 | 0.622 | -0.054 |
| spade | 0.765 | 0.882 | 0.118 |
| stress | 0.378 | 0.378 | |
| yew | 0.792 | 0.792 | |
| All | 0.649 | 0.669 | 0.020 |

Table 16: Precision Results Using ME to Disambiguate Words, With and Without Domains (recall and precision values are equal)

We obtained important conclusions about the relevance of domain information for each word. In general, the larger improvements appear for those words having well-differentiated domains (*spade*, *authority*). Conversely, the word *stress* with most senses belonging to the *FACTOTUM* domain do not improves at all. For example, for *spade*, *art* and *authority* (with an accuracy improvement over 10%) domain data seems to be an important source of knowledge with information that is not captured by other types of features. For those words for which precision decrease up to 6.5%, domain information is confusing. Three reasons can be exposed in order to explain this behavior: there is not a clear domain in the examples or they do not represent correctly the context, domains do not differentiate appropriately the senses, or the number of training examples is too low to perform a valid assessment. A cross-validation testing, if more examples were available, could be appropriate to perform a domain tuning for each word in order to determine which words must use this preprocess and which not.

Nevertheless, the experiment empirically demonstrates that a knowledge-based method, such as the domain heuristic, can be integrated successfully into a corpus-based system, such as maximum entropy, to obtain a small improvement.

## 4.3 Combining Results with(in) a Voting System

In previous sections, we have demonstrated that it is possible to integrate two different WSD approaches. In this section we evaluate the performance when combining a knowledge-based system, such as specification marks, and a corpus-based system, such as maximum entropy, in a simple voting schema.

In the two previous experiments we attempted to provide more information by pre-disambiguating the data. Here, we use both methods in parallel and then we combine their classifications in a voting system, both for Senseval-2 Spanish and English lexical–sample tasks.

### 4.3.1 Senseval-2 Spanish lexical–sample task

**vME+SM** is an enrichment of **vME**: we added the SM classifier to the combination of the three ME systems in **vME** (see Section 3.3). The results on the Spanish lexical–sample task from Senseval-2 are shown in Table 17. Because it only works with nouns, **vME+SM** improves accuracy for them only, but obtains the same score as JHU(R) while the overall score reaches the second place.

| ALL | | Nouns | |
|---|---|---|---|
| 0.713 | jhu(R) | 0.702 | jhu(R) |
| 0.684 | **vME+SM** | 0.702 | **vME+SM** |
| 0.682 | jhu | 0.683 | **MEbfs.pos** |
| 0.677 | **MEbfs.pos** | 0.681 | jhu |
| 0.676 | **vME** | 0.678 | **vME** |
| 0.670 | css244 | 0.661 | **MEbfs** |
| 0.667 | **MEbfs** | 0.652 | css244 |
| 0.658 | **MEfix** | 0.646 | **MEfix** |
| 0.627 | umd-sst | 0.621 | duluth 8 |
| 0.617 | duluth 8 | 0.612 | duluth Z |
| 0.610 | duluth 10 | 0.611 | duluth 10 |
| 0.595 | duluth Z | 0.603 | umd-sst |
| 0.595 | duluth 7 | 0.592 | duluth 6 |
| 0.582 | duluth 6 | 0.590 | duluth 7 |
| 0.578 | duluth X | 0.586 | duluth X |
| 0.560 | duluth 9 | 0.557 | duluth 9 |
| 0.548 | ua | 0.514 | duluth Y |
| 0.524 | duluth Y | 0.464 | ua |

Table 17: **vME+SM** in the Spanish lexical–sample task of Senseval-2

These results show that methods like SM and ME can be combined in order to achieve good disambiguation results. Our results are in line with those of Pedersen (2002), which also presents a comparative evaluation between the systems that participated in the Spanish and English lexical-sample tasks of Senseval-2. Their focus is on pair comparisons between systems to assess the degree to which they agree, and on measuring the difficulty of the test instances included in these tasks. If several systems are largely in agreement, then there is little benefit in combining them since they are redundant and they will simply reinforce each other. However, if some systems disambiguate instances that others do not, then the systems are complementary and it may be possible to combine them to take advantage of the different strengths of each system to improve overall accuracy.

The results for nouns (only applying SM), shown in Table 18, indicate that SM has a low level of agreement with all the other methods. However, the measure of optimal combination is quite high, reaching 89% (1.00–0.11) for the pairing of SM and JHU. In fact, all seven of the other methods achieved their highest optimal combination value when paired with the SM method.

| System pair for nouns | Both OK[a] | One OK [b] | Zero OK [c] | Kappa [d] |
|---|---|---|---|---|
| SM and JHU | 0.29 | 0.32 | 0.11 | 0.06 |
| SM and Duluth7 | 0.27 | 0.34 | 0.12 | 0.03 |
| SM and DuluthY | 0.25 | 0.35 | 0.12 | 0.01 |
| SM and Duluth8 | 0.28 | 0.32 | 0.13 | 0.08 |
| SM and Cs224 | 0.28 | 0.32 | 0.13 | 0.09 |
| SM and Umcp | 0.26 | 0.33 | 0.14 | 0.06 |
| SM and Duluth9 | 0.26 | 0.31 | 0.16 | 0.14 |

Table 18: Optimal combination between the systems that participated in the Spanish lexical–sample tasks of Senseval-2

---

a. Percentage of instances where both systems answers were correct.
b. Percentage of instances where only one answer is correct.
c. Percentage of instances where none of both answers is correct.
d. The kappa statistic (Cohen, 1960) is a measure of agreement between multiple systems (or judges) that is scaled by the agreement that would be expected just by chance. A value of 1.00 suggests complete agreement, while 0.00 indicates pure chance agreement.

This combination of circumstances suggests that SM, being a knowledge-based method, is fundamentally different from the others (i.e., corpus-based) methods, and is able to disambiguate a certain set of instances where the other methods fail. In fact, SM is different in that it is the only method that uses the structure of WordNet.

### 4.3.2 Senseval-2 English lexical–sample task

The same experiment was done on Senseval-2 English lexical–sample task data and the results are shown in Table 19. The details of how the different systems were built can be consulted in Section 3.2

Again, we can see in Table 19 that BFS per POS is better than per word, mainly because the same reasons explained in Section 3.3.

Nevertheless, the improvement on nouns by using the **vME+SM** system is not as high as for the Spanish data. The differences between both corpora have a significant relevance about the precision values that can be obtained. For example, the English data includes multi-words and the sense inventory is extracted from WordNet, while the Spanish data is smaller and a dictionary was built for the task specifically, having a smaller polysemy degree.

The results of **vME+SM** are comparable to the systems presented at Senseval-2 where the best system (Johns Hopkins University) reported 64.2% precision (68.2%, 58.5% and 73.9% for nouns, verbs and adjectives, respectively).

Comparing these results with those obtained in section 4.2, we also see that using a voting system with the best feature selection for ME and Specification Marks **vME+SM**, and using a non–optimized ME with the relevant domain heuristic, we obtain very similar performance. That is, it seems that we obtain comparable performance combining different

|         | All   | Nouns | Verbs | Adjectives |
|---------|-------|-------|-------|------------|
| MEfix   | 0.601 | 0.656 | 0.519 | 0.696      |
| MEbfs   | 0.606 | 0.658 | 0.519 | 0.696      |
| MEbfs.pos | 0.609 | 0.664 | 0.519 | 0.696    |
| vME+SM  | 0.619 | 0.667 | 0.535 | 0.707      |

**MEfix**: *0mcbWsdrvK3* for all words
**MEbfs**: each word with its
      best feature selection
**MEbfs.pos**: *0Wsrdm* for nouns,
      *0sbcprdmK10* for verbs,
      and *0mcbWsdrvK3* for adjectives
**vME+SM**: majority voting between MEfix,
      MEbfs.pos, MEbfs, and Specification Marks

Table 19: Precision Results Using Best Feature Selection for ME and Specification Marks on Senseval-2 English lexical–sample task data

classifiers resulting from a feature selection process or using a richer set of features (adding the domain features) with much less computational overhead.

This analysis of the results from the Senseval-2 English and Spanish lexical–sample tasks demonstrates that knowledge-based and corpus-based WSD systems can cooperate and can be combined to obtain improved WSD systems. The results empirically demonstrate that the combination of both approaches outperforms each of them individually, demonstrating that both approaches could be considered complementary.

## 5. Conclusions

The main hypothesis of this work is that WSD requires different kinds of knowledge sources (linguistic information, statistical information, structural information, etc.) and techniques. The aim of this paper was to explore some methods of collaboration between complementary knowledge-based and corpus-based WSD methods. Two complementary methods have been presented: specification marks (SM) and maximum entropy (ME). Individually, both have benefits and drawbacks. We have shown that both methods can collaborate to obtain better results on WSD.

In order to demonstrate our hypothesis, three different schemes for combining both approaches have been presented. We have presented different mechanisms of combining information sources around knowledge-based and corpus-based WSD methods. We have also shown that the combination of both approaches outperforms each of the methods individually, demonstrating that both approaches could be considered complementary. Finally, we have shown that a knowledge-based method can help a corpus-based method to better perform the disambiguation process, and vice versa.

In order to help the specification marks method, ME disambiguates some nouns in the context of the target word. ME selects these nouns by means of a previous analysis of training data in order to identify which ones seem to be highly accurately disambiguated.

This preprocess fixes some nouns reducing the search space of the knowledge-based method. In turn, ME is helped by SM by providing domain information of nouns in the contexts. This information is incorporated into the learning process in the form of features.

By comparing the accuracy of both methods, with and without the contribution of the other, it was demonstrated that such combining schemes of WSD methods are possible and successful.

Finally, we presented a voting system for nouns that included four classifiers, three of them based on ME, and one of them based on SM. This cooperation scheme obtained the best score for nouns when compared with the systems submitted to the SENSEVAL-2 Spanish lexical–sample task and comparable results to those submitted to the SENSEVAL-2 English lexical–sample task.

We are presently studying possible improvements in the collaboration between these methods, both by extending the information that the two methods provide to each other and by taking advantage of the merits of each one.

## Acknowledgments

The authors wish to thank the anonymous reviewers of the *Journal of Artificial Intelligence Research* and *COLING 2002, the 19th International Conference on Computational Linguistics*, for helpful comments on earlier drafts of the paper. An earlier paper (Suárez & Palomar, 2002b) about the corpus-based method (subsection 3.2) was presented at *COLING 2002*.

This research has been partially funded by the Spanish Government under project CICyT number TIC2000-0664-C02-02 and PROFIT number FIT-340100-2004-14 and the Valencia Government under project number GV04B-276 and the EU funded project MEANING (IST-2001-34460).